\documentclass[10pt,twocolumn,letterpaper]{article}

\usepackage{cvpr}              

\usepackage{times}
\usepackage{epsfig}
\usepackage{graphicx}
\usepackage{amsmath}
\usepackage{amssymb}
\usepackage{color}
\usepackage{bm}
\usepackage{cuted}
\usepackage{capt-of}
\usepackage{overpic}
\usepackage{verbatim}
\usepackage{multirow}
\usepackage{booktabs}
\usepackage{lipsum}

\usepackage[pagebackref,breaklinks,colorlinks]{hyperref}

\usepackage[capitalize]{cleveref}
\crefname{section}{Sec.}{Secs.}
\Crefname{section}{Section}{Sections}
\Crefname{table}{Table}{Tables}
\crefname{table}{Tab.}{Tabs.}

\def\cvprPaperID{6407} 
\def\confName{CVPR}
\def\confYear{2022}

\begin{document}

\title{Neural Points: Point Cloud Representation with Neural Fields for Arbitrary Upsampling}

\vspace{-10mm}

\vspace{-19mm}
\author{
	{\large Wanquan Feng\textsuperscript{1} \quad Jin Li\textsuperscript{2} \quad Hongrui Cai\textsuperscript{1}  \quad Xiaonan Luo\textsuperscript{2} \quad Juyong Zhang\textsuperscript{1}\thanks{Corresponding author} }
	\\
	{\normalsize \textsuperscript{1}University of Science and Technology of China \quad \textsuperscript{2}Guilin University Of Electronic Technology}
	\\ {\tt\footnotesize lcfwq@mail.ustc.edu.cn, 20032201014@mails.guet.edu.cn, hrcai@mail.ustc.edu.cn,} 
    \\ {\tt\footnotesize luoxn@guet.edu.cn, juyong@ustc.edu.cn }
}
\maketitle

\vspace*{-8mm}

\begin{abstract}
\vspace*{-4mm}
In this paper, we propose \emph{Neural Points}, a novel point cloud representation \textcolor{black}{and apply it to the arbitrary-factored upsampling task}. Different from traditional point cloud representation where each point only represents a position or a local plane in the 3D space, each point in Neural Points represents a local continuous geometric shape via neural fields. Therefore, Neural Points contain more shape information and thus have a stronger representation ability. Neural Points is trained with surface containing rich geometric details, such that the trained model has enough expression ability for various shapes. Specifically, we extract deep local features on the points and construct neural fields through the local isomorphism between the 2D parametric domain and the 3D local patch. In the final, local neural fields are integrated together to form the global surface. Experimental results show that Neural Points has powerful representation ability and demonstrate excellent robustness and generalization ability. With Neural Points, we can resample point cloud with arbitrary resolutions, and it \textcolor{black}{outperforms the state-of-the-art point cloud upsampling methods. Code
is available at \href{https://github.com/WanquanF/NeuralPoints}{https://github.com/WanquanF/NeuralPoints}.}
\end{abstract}


\vspace*{-7mm}

\section{Introduction}
\label{Introduction}

\vspace*{-2mm}

Point cloud, which is the most fundamental and popular representation of 3D scenes, has been widely used in many applications like 3D reconstruction~\cite{HoppeDDMS92, Newcombe2012KinectFusion, Kazhdan2013Screened, Riegler2018OctNetFusion}, virtual/augmented reality~\cite{HeldGCA12,santana2017multimodal} and autonomous driving~\cite{WangCGHCW19,lang2019pointpillars}. In the traditional representation of point cloud, each point only represents a position in the 3D space, and it can be further extended to represent a local plane if its normal vector is assigned. Therefore, the representation ability of point cloud is still limited by its resolution. Although point cloud upsampling methods~\cite{YuLFCH18,WangW0CS19,LiLFCH19,LiLHF21} have been proposed to improve its representation ability, their strategy is still a ``discrete-to-discrete'' manner and can not overcome the limitation of current point cloud representation.

In this work, we propose \emph{Neural Points}, a novel point cloud representation, which significantly improves the representation ability \textcolor{black}{and can be naturally applied to the upsampling task}. Different from traditional representation, each point of Neural Points encodes a local surface patch represented via neural fields. Specifically, each point-wise surface patch is represented as a local isomorphism between the 2D parametric domain and the 3D local surface patch, and the isomorphism is implicitly represented via neural fields. Thanks to its continuous nature, free of the limitation of finite resolution, powerful representation ability to \textcolor{black}{involve more shape information}, neural fields enable Neural Points several advantages over traditional discrete point cloud representation. Meanwhile, the trained model of neural fields is shared for all local patches, and thus the storage overhead of Neural Points is quite small.


Unlike some existing methods which represent the whole surface model via one neural field~\cite{park2019deepsdf,yariv2020multiview}, both the neural fields and the point features passed into Neural Points are all local. With this design, our Neural Points representation shows several advantages including excellent ability to express details, strong generalization ability, and low need for training data. Specifically, we employ local neural fields to construct the continuous bijective mapping between the 2D parametric domain and the 3D local surface patch, based on the fact that 3D local patch on the 2D manifold is isomorphic to a 2D simply connected disk. With the neural field, each 3D local patch can be viewed as a parametric surface defined in a 2D parametric domain. To this end, we employ an encoder to extract point-wise local features and take the local feature as a part of the input passed into the neural field network. With the local shape information, the neural field can represent the surface patch well, and the trained network model is shared by all local patches of all 3D models. In the final, we design an integration strategy to integrate all the local neural fields together to form the final global shape. The model of Neural Points is trained with geometric surface in high resolution with rich geometric details, such that \textcolor{black}{the trained Neural Points model can represent rich geometric shapes}.

\textcolor{black}{As for the point cloud upsampling task, its target is to predict the upsampled point cloud with higher resolution from the given low-resolution input such that it can better capture the geometric features of the underlying continuous surface.} Existing point upsampling methods~\cite{YuLFCH18,WangW0CS19,LiLFCH19,LiLHF21,LuoTZWY21,shuquanMetaPU21,ZhaoHX21,QianHKH21} usually employ a ``discrete-to-discrete'' manner. Dissimilarly, we convert the input point cloud representation to continuous Neural Points, and then apply arbitrary-factored sampling.

The experimental results show that Neural Points can produce high-quality continuous surfaces and overcome the limitation of point resolution well. For the upsampling task, our strategy \textcolor{black}{outperforms the state-of-the-art methods}. In summary, the contributions of this work include: 
\vspace*{-2.5mm}  
\begin{itemize}
    \item We propose Neural Points, a novel point cloud representation \textcolor{black}{that can be naturally applied to the upsampling task}, with low storage overhead.   
    \vspace*{-2.5mm}  
    \item We employ the neural implicit functions along with the deep local features of surface patches to represent the local neural fields, and design an integration strategy to form the final global shape. 
    \vspace*{-2.5mm}  
    \item Evaluation results demonstrate that the Neural Points representation has excellent robustness and generalization ability for various inputs on the upsampling task. 
    \vspace*{-0.5mm}  
\end{itemize}
\vspace{-1mm}
\section{Related Works}
\label{Related}
\begin{figure*}[t]
\centering
\includegraphics[width=1.0\textwidth]{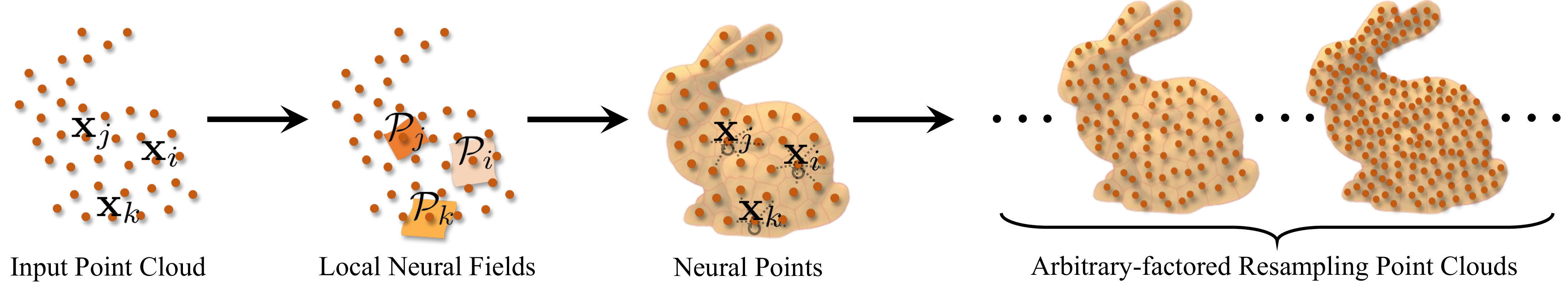}
\vspace{-7mm}
\caption{Algorithm pipeline. For the input point cloud, a discrete point-wise local patch is represented via local continuous neural fields, and the global continuous Neural Points surface is constructed by integrating all the local neural fields. Arbitrary resolutions of point cloud can be generated by sampling on the constructed continuous Neural Points surface.}
\label{fig:pipeline}
\vspace{-5mm}
\end{figure*} 

\vspace{-1mm}
{\noindent \bf Neural Implicit Function.} Neural implicit representation has recently shown promising advantages in some applications over discrete scene representations, such as point clouds, meshes and voxel grids, owing to its continuous nature and free of the limitation of finite resolution. It has also achieved great performance in many applications including 3D shape representation~\cite{park2019deepsdf,mescheder2019occupancy,chen2019learning}, novel view synthesis~\cite{sitzmann2019scene,MildenhallSTBRN20} and multi-view reconstruction~\cite{niemeyer2020differentiable,yariv2020multiview}.

For 3D shape representation, DeepSDF~\cite{park2019deepsdf} and ONet~\cite{mescheder2019occupancy} proposed to use neural implicit function to predict the signed distance and occupancy probability, respectively. However, these conventional frameworks are weak in representing complex shapes. To alleviate this issue, some recent works focus on geometric detail preservation and local shape reduction. IF-Nets~\cite{chibane2020implicit} extracted learnable multi-scale features, which encodes local and global properties of 3D shape. LGCL~\cite{yao20213d} represented 3D shape as zero-surfaces with local latent codes, leading each local SDF responsible for a part of the shape. Due to local feature learning, these methods represent higher-quality details than common methods which use a single vector to encode a shape. In our method, we also devise a novel and effective strategy to extract deep local features from point cloud.

Neural implicit functions have also been used to represent 2D images. LIIF~\cite{ChenL021} took a pixel coordinate and its surrounding feature as input to predict the corresponding RGB value. It demonstrates excellent super-resolution effect, building a bridge between discrete and continuous representation in the 2D domain. Different from regular 2D image, irregular point cloud is a discrete representation of 3D surface embedded in 2D manifold. To handle the irregular representation, we construct the mapping between the irregular local surface patch and the regular 2D domain.

\vspace{-0.2mm}
{\noindent \bf Optimization Based Point Upsampling.} In optimization based point upsampling methods, shape priors like global structures and local smoothness are formulated as objective energy to constrain the results. Alexa \emph{et al.}~\cite{AlexaBCFLS03} employed Voronoi diagram and interpolated points at vertices of the Voronoi diagram in local tangent space as upsampled points. Then Lipman \emph{et al.}~\cite{LipmanCLT07} proposed a parameter-free method based on the locally optimal projection operator (LOP) for point sampling and surface reconstruction. Later, weighted LOP~\cite{HuangLZAC09} conducted an iterative normal estimation process to consolidate the upsampled point. EAR~\cite{0004WGCAZ13} sampled points away from edges and progressively approaching edges and corners. Then a point-set consolidation method~\cite{Wu0GZC15} proposed to fill large holes.

\vspace{-0.2mm}
{\noindent \bf Learning Based Point Upsampling.} Considering that the point upsampling task is an ill-posed problem, learning priors from dataset is a natural way to tackle it. In recent years, point-based network structures~\cite{QiSMG17,QiYSG17,WangSLSBS19,LiBSWDC18,KomarichevZH19,feng2021recurrent} have been successfully employed to solve point cloud related tasks and achieved state-of-the-art performance. PU-Net~\cite{YuLFCH18} was the first point upsampling network. 
MPU~\cite{WangW0CS19} was a progressive upsampling method that achieved great performance for large upsampling factors. PU-GAN~\cite{LiLFCH19} was the first the GAN~\cite{GoodfellowPMXWOCB14} based structure to synthesize uniformly distributed points. While the above mentioned works are mainly focused on the network design, PUGeo-Net~\cite{QianHK020} \textcolor{black}{and MAFU~\cite{QianHKH21}} utilized local differential geometry constraints to improve the upsampling results. \textcolor{black}{Meta-PU~\cite{shuquanMetaPU21} firstly upsampled to a large resolution and then downsampled to a target resolution}. PU-GCN~\cite{QianALTG21} improved the performance by proposing a novel NodeShuffle module. Dis-PU~\cite{LiLHF21} employed a global refinement module at the end of the pipeline instead of using a single upsampling module. However, all the above methods adopt the strategy to directly predict a denser point cloud from the input sparse point cloud. Unlike all these methods, we utilize neural fields to represent the high-resolution surface, which has a more powerful representation ability and can then be sampled in arbitrary resolutions.

\vspace{-1mm}

\section{Neural Points}
\label{Algorithm}

\vspace{-2mm}
\subsection{Overview: Point based Representation}
\label{overview}
Point cloud $\mathcal{X}=\{\mathbf{x}_i \in \mathbb{R}^{3}\}_{i=1}^{I}$ is the discrete representation of its underlying continuous surface $\mathcal{S}$. For the traditional point cloud representation, where each point only represents a 3D position, its representation ability totally depends on its resolution. Therefore, one direct strategy to improve its representation ability is to do upsampling: 
\vspace{-2mm}
\begin{equation}
    \begin{split}
        \mathcal{S} \xrightarrow{\textrm{Discretize}} \{\mathbf{x}_i\}_{i=1}^{I}  \xrightarrow{\textrm{Upsample}} \{\mathbf{x}_i^{r}\}_{i,r=1}^{I,R} \subset \mathcal{S},
    \label{improve_ability_discrete}
\vspace{-6mm}
    \end{split}
\end{equation}
which was studied in the point cloud upsampling works~\cite{YuLFCH18,LiLHF21}. However, the upsampling manner in Eq.~\eqref{improve_ability_discrete} is discrete-to-discrete, where the upsampled result is still discrete and limited by the resolution.

In this work, we propose \emph{Neural Points}, a novel point cloud representation \textcolor{black}{for arbitrary upsampling}, which has better representation ability than the traditional point cloud. Neural Points representation employs the discrete-to-continuous strategy, which is totally different from the form of Eq.~\eqref{improve_ability_discrete}. Given the input point cloud $\mathcal{X}=\{\mathbf{x}_i\}_{i=1}^{I}$, we describe the underlying continuous surface 
\vspace{-2mm}
\begin{equation}
    \begin{split}
        \mathcal{S} \xrightarrow{\textrm{Discretize}} \{\mathbf{x}_i\}_{i=1}^{I}  \xrightarrow{\textrm{Neural Points}} \mathcal{S}^{'} \approx \mathcal{S},
    \label{improve_ability_np}
    \end{split}
\vspace{-6mm}
\end{equation}
where $\mathcal{S}^{'}$ is a continuous surface represented by Neural Points. With the form in Eq.~\eqref{improve_ability_np}, our proposed framework can overcome the limitation of point cloud resolution and achieve arbitrary-factored point cloud sampling on $\mathcal{S}^{'}$. In the following, we will describe the algorithm details of our Neural Points representation.

\begin{figure}[t]
\centering
\includegraphics[width=1.0\columnwidth]{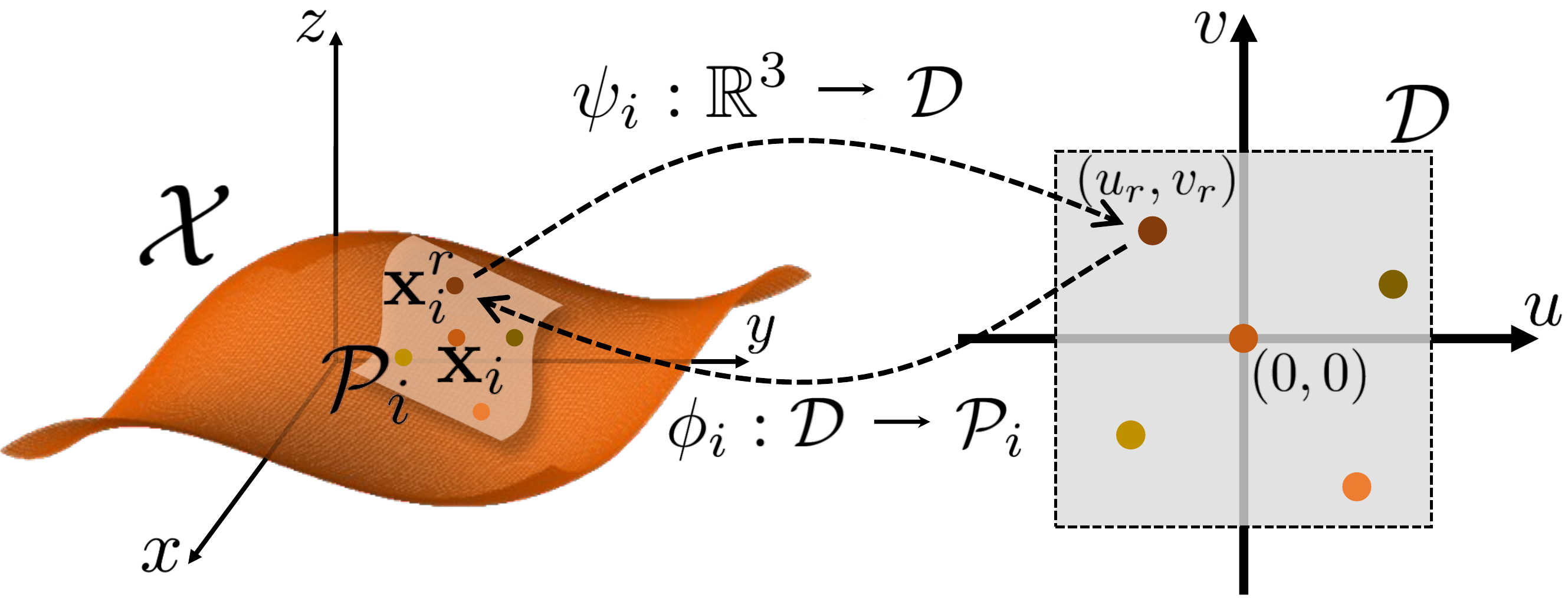}
\vspace{-8mm}
\caption{Local neural field and the bijective mapping function between isomorphic 3D and 2D domain. $\mathcal{P}_i$ is the 3D local surface patch around the center point $\mathbf{x}_i$, and $\mathcal{D}$ is the 2D parametric domain. $\phi_i$ and $\psi_i$ are the mappings that constructs the correspondence between $(u_r,v_r) \in \mathcal{D}$ and $\mathbf{x}_i^r \in \mathcal{P}_i$.}
\label{fig:local_para}
\vspace{-5mm}
\end{figure} 

\subsection{Pipeline}
\label{pipeline}

The whole pipeline is given in Fig.~\ref{fig:pipeline}. (1) Given the input point cloud, we first construct local neural fields for each local patch, which is based on local parameterization; (2) The local neural fields are integrated together to form the global shape; (3) With the constructed continuous neural representation, we can resample an arbitrary number of points. In the following, we introduce the details of each component.

{\noindent \bf Local Neural Fields.} 
We employ local neural fields to describe the underlying continuous surface. The input point cloud is still denoted as $\mathcal{X} = \{\mathbf{x}_i\}_{i=1}^{I}$. Like in~\cite{QianHK020}, we take $\{\mathbf{x}_i\}_{i=1}^{I} $ as the center points to divide the surface into overlapping local patches $\{\mathcal{P}_i\}_{i=1}^{I} \subset \mathbb{R}^3$. At each point $\mathbf{x}_i$, the 3D local patch $\mathcal{P}_i$ is isomorphic to the 2D parametric domain $\mathcal{D} \subset \mathbb{R}^2$ (we use $\mathcal{D}=[-1,1]^2$ in our work), which means that we can construct a bijective mapping between them: 
\vspace{-1mm}
        $$\phi_{i}: \mathcal{D}  \rightarrow \mathcal{P}_i,\quad \psi_{i}: \mathbb{R}^3  \rightarrow \mathcal{D},$$
where $\psi_i|_{\mathcal{P}_i}=\phi_i^{-1}$. The illustration of the neural field is shown in Fig.~\ref{fig:local_para}. Given any 2D sampling point $(u_r,v_r) \in \mathcal{D}$, we can compute $\mathbf{x}_i^r = \phi_{i}(u_r,v_r) \in \mathcal{P}_i$ as its corresponding 3D sampling point. Similarly, for any 3D point $\mathbf{x}_i^r \in \mathcal{P}_i$, we can compute its corresponding 2D coordinate $(u_r,v_r) = \psi_i(\mathbf{x}_i^r) \in \mathcal{D}$. Furthermore, we can compute the point normal $\mathbf{n}_{\mathbf{x}_i^r}$ at point $\mathbf{x}_i^r$ via:
\vspace{-1mm}
        $$\mathbf{n}_{\mathbf{x}_i^r} =  \nabla_u \phi_i \times \nabla_v \phi_i ,$$
and then normalize it to unit length.

{\noindent \bf Neural Fields Integration.} 
Although the local neural fields $\{\phi_i\}_{i=1}^{I}$ can construct correspondence between 2D parametric domain and 3D local patches, different local neural fields are defined in different local coordinate systems. As a result, we need to integrate them together to obtain a globally continuous neural field based surface.

For $\forall \mathbf{x} \in \mathbb{R}^3$, we aim to pull $\mathbf{x}$ onto the neural fields based surface. We find the nearest center points in $\{\mathbf{x}_i\}_{i=1}^{I}$ to $\mathbf{x}$ and denote the set of indexes of the neighbour points as $\mathcal{N}(\mathbf{x})$. For $\forall k \in \mathcal{N}(\mathbf{x})$, a combination weight is computed according to the distance from $\mathbf{x}$ to $\mathbf{x}_k$:
\begin{equation}
    \begin{split}
    w_k = e^{- \alpha_1 \|\mathbf{x}-\mathbf{x}_k\|_2^2}, \forall k \in \mathcal{N}(\mathbf{x}).
        \label{skining_weight}
    \end{split}
\end{equation}
With $w_k$, the point on the neural fields based surface can be computed by:
\begin{equation}
    \begin{split}
     \rho(\mathbf{x})= (\sum\limits_{k \in \mathcal{N}(\mathbf{x})}w_k \cdot \phi_k \circ \psi_k (\mathbf{x}))/(\sum\limits_{k \in \mathcal{N}(\mathbf{x})}w_k), \forall \mathbf{x} \in \mathbb{R}^3.
        \label{upsampled_point_coor}
    \end{split}
\end{equation}
Specifically, the input point $\mathbf{x}$ is mapped to 2D parametric domain through the neighbour neural fields as $\psi_k (\mathbf{x}) \in \mathcal{D}$, and then map the 2D point back to 3D coordinate as $\phi_k \circ \psi_k (\mathbf{x}) \in \mathcal{P}_k$. In this way, a neural fields based continuous surface is constructed.

The normal at point $\rho(\mathbf{x})$ is computed as:
\begin{equation}
    \begin{split}
        \mathbf{n}_{\rho(\mathbf{x})} = (\sum\limits_{k \in \mathcal{N}(\mathbf{x})}w_k \cdot \mathbf{n}_{\phi_k \circ \psi_k (\mathbf{x})})/(\sum\limits_{k \in \mathcal{N}(\mathbf{x})}w_k), \forall \mathbf{x} \in \mathbb{R}^3,
        \label{point_normal_pho}
    \end{split}
\end{equation}
and the combination vector is normalized to unit length. Note that the orientations of $\{\mathbf{n}_{\phi_k \circ \psi_k (\mathbf{x})}\}_{k \in \mathcal{N}(\mathbf{x})}$ are adjusted to be along the same direction before computing the weighted sum.

{\noindent \bf Point Cloud Sampling.}
We can also resample the point cloud from the neural fields represented continuous surface. Considering that the point cloud sampling process is essentially applied on the 2D manifold, we start the sampling operation in the 2D parametric domain $\mathcal{D}$.

We sample points uniformly in the 2D parametric domain $\mathcal{D}$, and the 2D sampled points are mapped onto the 3D local patch. Specifically, for center point $\mathbf{x}_i$, we uniformly sample $R$ points $\{(u_r,v_r) \in \mathcal{D} \}_{r=1}^{R}$ and then map them to 3D as $\{\mathbf{x}_i^r \in \mathcal{P}_i \}_{r=1}^{R}$. For the whole input point cloud, we obtain the union of sampled points from all patches:
   $$ \mathcal{X}_R = \bigcup\limits_{i=1}^I (\{\mathbf{x}_i^r \in \mathcal{P}_i \}_{r=1}^{R}) = \{\mathbf{x}_i^r \in \mathcal{P}_i \}_{i,r=1}^{I,R}.$$
Then we can uniformly sample $J$ points from $\mathcal{X}_R$:
   $$ \mathcal{Y}^{*}=\{\mathbf{y}_j^{*} \in \bigcup\limits_{i=1}^I \mathcal{P}_i \}_{j=1}^{J},$$
where the value of $J$ is arbitrary in the training and inferring stages. Then we pull $\mathbf{y}_j^{*}$ onto the Neural Points surface with Eq.~\eqref{upsampled_point_coor} as:
  $$  \mathcal{Y}=\{\mathbf{y}_j = \rho(\mathbf{y}_j^{*}) \}_{j=1}^J.$$

\subsection{Network Structure}
\label{network_struture}
In this part, we introduce the network structures, including the local feature extraction and the local neural fields.

{\noindent \bf Local Feature Extraction.}
As discussed above, for each center point $\mathbf{x}_i$, we extract a local feature as a part of the input into the neural field $\phi_i$, which we denote as $\mathbf{f}_i$. For each $\mathbf{x}_i$, we first extract its neighbour points $\{\mathbf{x}_k\}_{k \in \mathcal{N}(\mathbf{x}_i)}$ and decentralize to $\{\mathbf{x}_k-\mathbf{x}_i\}_{k \in \mathcal{N}(\mathbf{x}_i)}$. Then we apply the DGCNN\cite{WangSLSBS19} backbone to extract feature on the point set with $|\mathcal{N}(\mathbf{x}_i)|$ points. Specifically, we employ several EdgeConv layers with dynamic graph update. The features of each layer are concatenated and then passed into another EdgeConv layer and max-pooling layer to get $\mathbf{f}_i^{*}$. For each $i$, we concatenate $\mathbf{f}_i^{*}$ with the local pooling of its neighbors to get the final local feature:
  $$  \mathbf{f}_i = \mathbf{f}_i^{*} \oplus MaxPool\{\mathbf{f}_k^{*}\}_{k \in \mathcal{N}(\mathbf{x}_i)},$$
where $\oplus$ denotes the concatenation operation.

{\noindent \bf Structure of Neural Fields.}
We employ the MLP-based network equipped with the ReLU activation layers to represent $\phi_i$. The implicit functions are shared by all patches of all 3D models, and they are denoted as $\Phi$. As we mentioned above, the input of $\Phi$ should contain the local shape information involved in $\mathbf{f}_i$ and the 2D parametric coordinates as the querying point.

For the 2D querying coordinates passed into $\phi_i$, we apply position encoding as the design in~\cite{MildenhallSTBRN20} and denote the position encoding function as $\gamma$. The position code is concatenated with the local feature as the input of the neural implicit function. Specifically, we can formulate $\Phi$ as:
    $$\Phi((u_r, v_r), \mathbf{f}_i, \mathbf{x}_i) = \mathbf{x}_i + \theta_{\Phi}(\gamma(u_r,v_r) \oplus \mathbf{f}_i) ,$$ 
where $\theta_{\Phi}$ is the MLP in $\Phi$. Naturally, for each $i$, we can formulate $\phi_i(\cdot) = \Phi(\cdot,\mathbf{f}_i,\mathbf{x}_i)$.

Another key function is $\psi_{i}$, who is the extension of $\phi_i^{-1}$. In our implementation, $\psi_i$ is defined as:
\begin{equation}
    \begin{split}
    \psi_i(\mathbf{x}) = \phi_i^{-1}(Proj(\mathbf{x}, \phi_i(\mathcal{D}))), \forall \mathbf{x} \in \mathbb{R}^3,
        \label{psi_formulation}
    \end{split}
\end{equation}
where we use $Proj(\mathbf{x}, \phi_i(\mathcal{D}))$ to denote the nearest point in $\phi_i(\mathcal{D})$ to $\mathbf{x}$. The advantage of this definition is that we can formulate $\phi_i \circ \psi_i$ as:
\begin{equation}
    \begin{split}
    \phi_i \circ \psi_i(\mathbf{x}) = Proj(\mathbf{x}, \phi_i(\mathcal{D})), \forall \mathbf{x} \in \mathbb{R}^3,
        \label{phi_psi_formulation}
    \end{split}
\end{equation}
which is used in Eq.~\eqref{upsampled_point_coor}.

At last, we explain our projection process in Eq.~\eqref{psi_formulation} and Eq.~\eqref{phi_psi_formulation}. In our implementation, we use the approximation for $\phi_i(\mathcal{D})$ as:
    $$\phi_i(\mathcal{D}) \approx \{\mathbf{x}_i^r\}_{r=1}^{R}.$$
Then, the projection operation should be formulated from a 3D point and a point set. For the convenience of description, we consider a 3D point denoted as $\mathbf{p}$ and a point set denoted as $\mathcal{Q}=\{\mathbf{q}_t\}_{t=1}^T$. We find the index set of neighbor points of $\mathbf{p}$ in $\mathcal{Q}$ and denote it as $\mathcal{N}(\mathbf{p};\mathcal{Q})$. The projection from $\mathbf{q}$ to $\mathcal{Q}$ is formulated as:
\begin{equation}
    \begin{split}
    Proj(\mathbf{p}, \mathcal{Q}) = (\sum\limits_{k \in \mathcal{N}(\mathbf{p};\mathcal{Q})}w_k \cdot \mathbf{q}_k)/(\sum\limits_{k \in \mathcal{N}(\mathbf{p};\mathcal{Q})}w_k),
        \label{projection}
    \end{split}
\end{equation}
where $w_k$ is computed as:
\begin{equation}
    \begin{split}
    w_k = e^{- \alpha_2 \|\mathbf{p}-\mathbf{q}_k\|_2^2}, \forall k \in \mathcal{N}(\mathbf{p};\mathcal{Q}).
        \label{skining_weight_2}
    \end{split}
\end{equation}
Similar to Eq.~\eqref{point_normal_pho}, its normal $\mathbf{n}_{Proj(\mathbf{p}, \mathcal{Q})}$ can be obtained.

\subsection{Loss Function}
\label{loss}
In this part, we introduce the loss terms, including the constraints on surface shape, point normal, and the integration quality. The total loss is:
        $$\mathcal{L} = \mathcal{L}_{\textrm{shape}} + \omega_{1} \cdot \mathcal{L}_{\textrm{nor}} + \omega_{2} \cdot \mathcal{L}_{\textrm{int}}.$$
Point clouds with higher resolution are used as ground truth in our current implementation, and other shape representations can also be used accordingly. We denote the ground truth point cloud as $\mathcal{Z}=\{\mathbf{z}_j\}_{l=1}^{L}$ and denote the normal of $\mathbf{z}_l$ as $\mathbf{n}_{\mathbf{z}_l}$. As described in Sec.~\ref{pipeline}, the output of the framework can be summarized as $\mathcal{X}_{R} = \{\mathbf{x}_i^r; \mathbf{n}_{\mathbf{x}_i^r}\}_{i,r=1}^{I,R}$ and $\mathcal{Y} = \{\mathbf{y}_j; \mathbf{n}_{\mathbf{y}_j}\}_{j=1}^{J}$.

The supervision is discrete, but $\mathcal{X}_{R}$ and $\mathcal{Y}$ should be viewed as arbitrary sampling from continuous surface. To supervise a continuous surface with the discrete supervision, we do not employ the Chamfer loss which is based on closest point searching, though it was widely used in the previous point upsampling works~\cite{QianALTG21,LiLHF21}. Instead, we employ the projection strategy described in Eq.~\eqref{projection}. Generally, for two point clouds $\mathcal{P}=\{\mathbf{p}_s\}_{s=1}^{S}$ and $\mathcal{Q}=\{\mathbf{q}_t\}_{t=1}^{T}$, we define their distance as:
      $$  d(\mathcal{P},\mathcal{Q}) = \frac{1}{S}\sum\limits_{s=1}^{S} \|\mathbf{p}_s-Proj(\mathbf{p}_s,\mathcal{Q})\|_2^2.$$
We also define their difference of point-wise normal as:
        $$d_{\mathbf{n}}(\mathcal{P},\mathcal{Q}) = \frac{1}{S}\sum\limits_{s=1}^{S} \|\mathbf{n}_{\mathbf{p}_s}-\mathbf{n}_{Proj(\mathbf{p}_s,\mathcal{Q})}\|_2^2.$$
To constrain $\mathcal{X}_{R}$ and $\mathcal{Y}$ to be close to $\mathcal{Z}$, the loss term is:
        $$\mathcal{L}_{\textrm{shape}} = d(\mathcal{X}_{R},\mathcal{Z}) +
        d(\mathcal{Z},\mathcal{X}_{R}) + d(\mathcal{Y},\mathcal{Z}) +
        d(\mathcal{Z},\mathcal{Y}).$$
To supervise the point-wise normal, the loss term is:
       $$ \mathcal{L}_{\textrm{nor}} = d_{\mathbf{n}}(\mathcal{X}_{R},\mathcal{Z}) + d_{\mathbf{n}}(\mathcal{Z},\mathcal{X}_{R}) + d_{\mathbf{n}}(\mathcal{Y},\mathcal{Z}) + d_{\mathbf{n}}(\mathcal{Z},\mathcal{Y}).$$
Furthermore, we also employ a loss term for the integration quality. The overlap shape from neighbour neural fields should be identical, and the constructed global neural points surface should cover the input surface. Based on these requirements, we design the loss term as:
        $$\mathcal{L}_{\textrm{int}} = \sum\limits_{j=1}^{J} \sum\limits_{k \in \mathcal{N}(\mathbf{y}_j;\mathcal{X})}\|\mathbf{y}_j-Proj(\mathbf{y}_j,\{\mathbf{x}_k^r\}_{r=1}^R)\|_2^2.$$

\section{Experiments}
\label{Experiments}

In this section, we give the implementation details, ablation studies, results, comparisons and the test on generalization and robustness.

\subsection{Implementation Details}
\label{implementation_details}

\noindent{\bf Dataset.}
We train and test our model on Sketchfab~\cite{sketchfab} dataset collected by PUGeo-Net~\cite{QianHK020}, which contains $90$ training and $13$ testing models with rich geometry features. We train all the comparison methods with the same dataset for fair comparison. Similar to other point cloud upsampling methods, we employ Poisson disk sampling~\cite{CorsiniCS12} algorithm to extract points on the models to obtain the input and ground truth of the upsampling algorithm. Specifically, we extract $10,000$ points as the whole input point cloud and extract $40,000$ and $160,000$ points as the whole ground truth point cloud for $4\times$ and $16\times$ experiments. We also follow the previous point upsampling works to extract some anchor points and the neighboring local parts (subsets of the whole point cloud) as the input of the network instead of passing the whole model into the network. For all, we choose $1,000$ anchors on each training model and $114$ anchors on each testing model. During training, the point number of the input point clouds is set as $256$. 

To test the effectiveness of our trained model, we not only test on the testing set of Sketchfab, but also test on more unseen datasets without retraining our network. We test all the methods on dataset collected by PU-GAN~\cite{LiLFCH19}, where the number of testing models is $27$. The shape of the testing models in this dataset is relatively simple, and thus we extract $2,000$ points on each model as the whole input point cloud. In addition to the synthetic data mentioned above, we further test on real captured data. We also evaluate our method on point clouds captured by depth sensor on iPhone X \textcolor{black}{and LiDAR data from KITTI~\cite{GeigerLSU13}}.

\noindent{\bf Experimental Setting.} 
All the inputs passed into the network are normalized into the 3D unit ball. In all our experiments, we set $R= \lfloor 4 \cdot J / I \rfloor$. In the local feature extracting network, the number of neighboring point set is $10$. For all the $Proj(.,.)$ operations, we set the number of neighboring points as $4$. For the loss terms, $\omega_1$ and $\omega_2$ are set as $0.01$ and $0.3$, respectively. $\alpha_1$ in Eq.~\eqref{skining_weight} and $\alpha_2$ in Eq.~\eqref{skining_weight_2} are set as $10^2$ and $10^3$ respectively to compute the exponential weights. We employ $5$ convolution layers in the DGCNN backbone and $3$ linear layers in the MLP of $\Phi$. We set the batch size as $8$ and train the network with $25,000$ iterations for all. The learning rate starts as $0.01$, and is multiplied by $0.5$ every $1,250$ iterations. For all the comparison methods, we train them with the same settings as our own method. The model is trained with PyTorch~\cite{PaszkeGMLBCKLGA19}. All the training and testing is conducted on a workstation with four 32G V100 GPUs, 32 Intel(R) Xeon(R) Silver 4110 CPU @ 2.10GHz, and 128GB of RAM. Our trained network model can be applied to all point clouds. The storage overhead of Neural Points is to store the pre-trained model, whose total size is only $2.53$MB ($1.35$MB for the local feature extracting network and $1.18$MB for the Neural field MLP network.).

\noindent{\bf Evaluation Metric.} Similar to recent point cloud upsampling works~\cite{QianALTG21,LiLHF21}, we employ Chamfer Distance (CD), Hausdorff Distance (HD), and Point-to-Surface (P2F) as the metrics. For all the metrics, the smaller the metric, the better the quality of the results.

\begin{figure*}[t]
\centering
\includegraphics[width=0.95\textwidth]{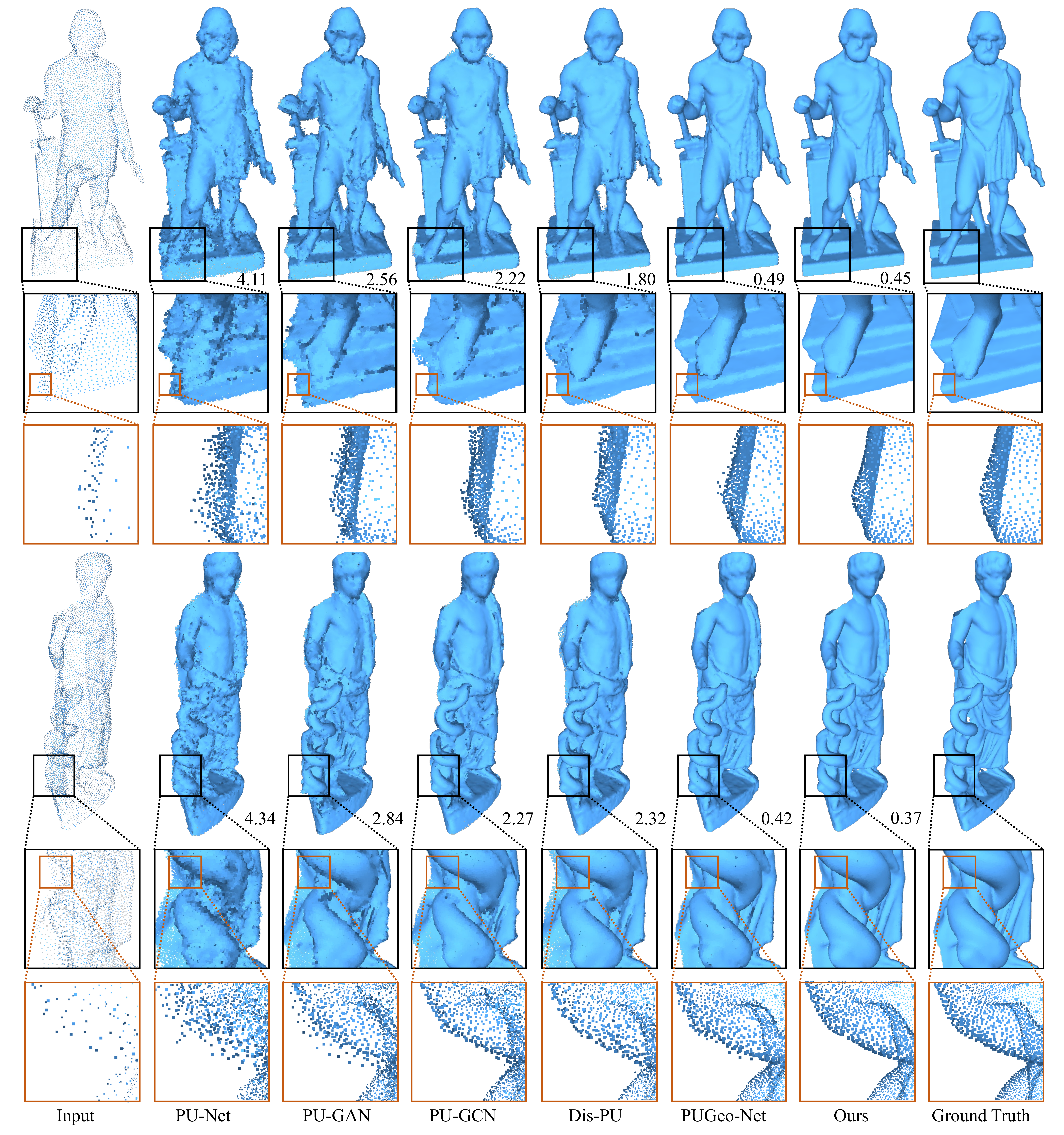}
\vspace{-4mm}
\caption{Results and comparison on the Sketchfab~\cite{sketchfab} dataset. The error metric CD ($\times 10^{-5}$) is also given in the bottom. For better visualization, we zoom-in some local parts of the results and choose the appropriate views to show the details.}
\label{fig:sketchfab}
\vspace{-5mm}
\end{figure*}

\begin{figure*}[t]
\centering
\includegraphics[width=1.0\textwidth]{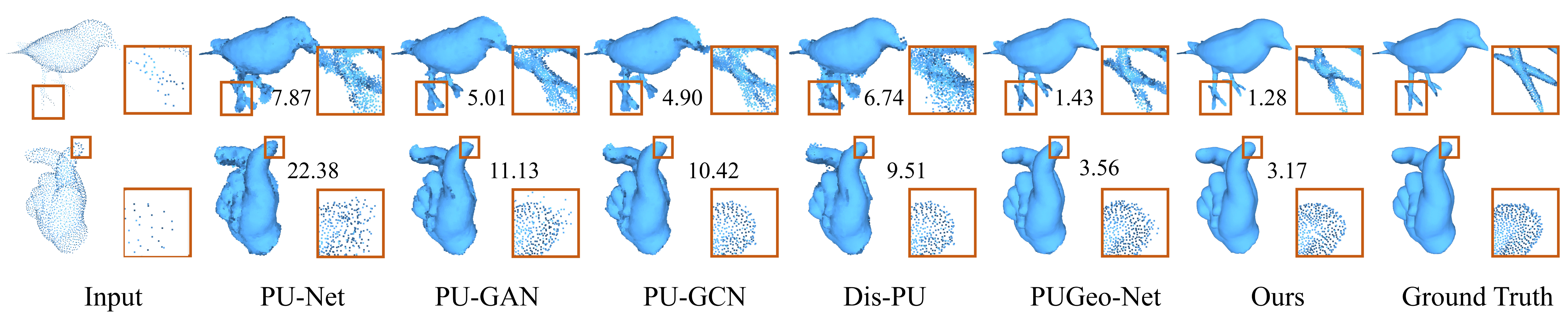}
\vspace{-8mm}
\caption{Results and comparison on the PU-GAN~\cite{LiLFCH19} dataset. The error metric CD ($\times 10^{-5}$) is also given in the bottom. Some local parts are displayed for better comparison.}
\label{fig:pugan}
\vspace{-5mm}
\end{figure*}

\vspace{-0.1mm}
\begin{figure}[h]
\centering
\includegraphics[width=0.47\textwidth]{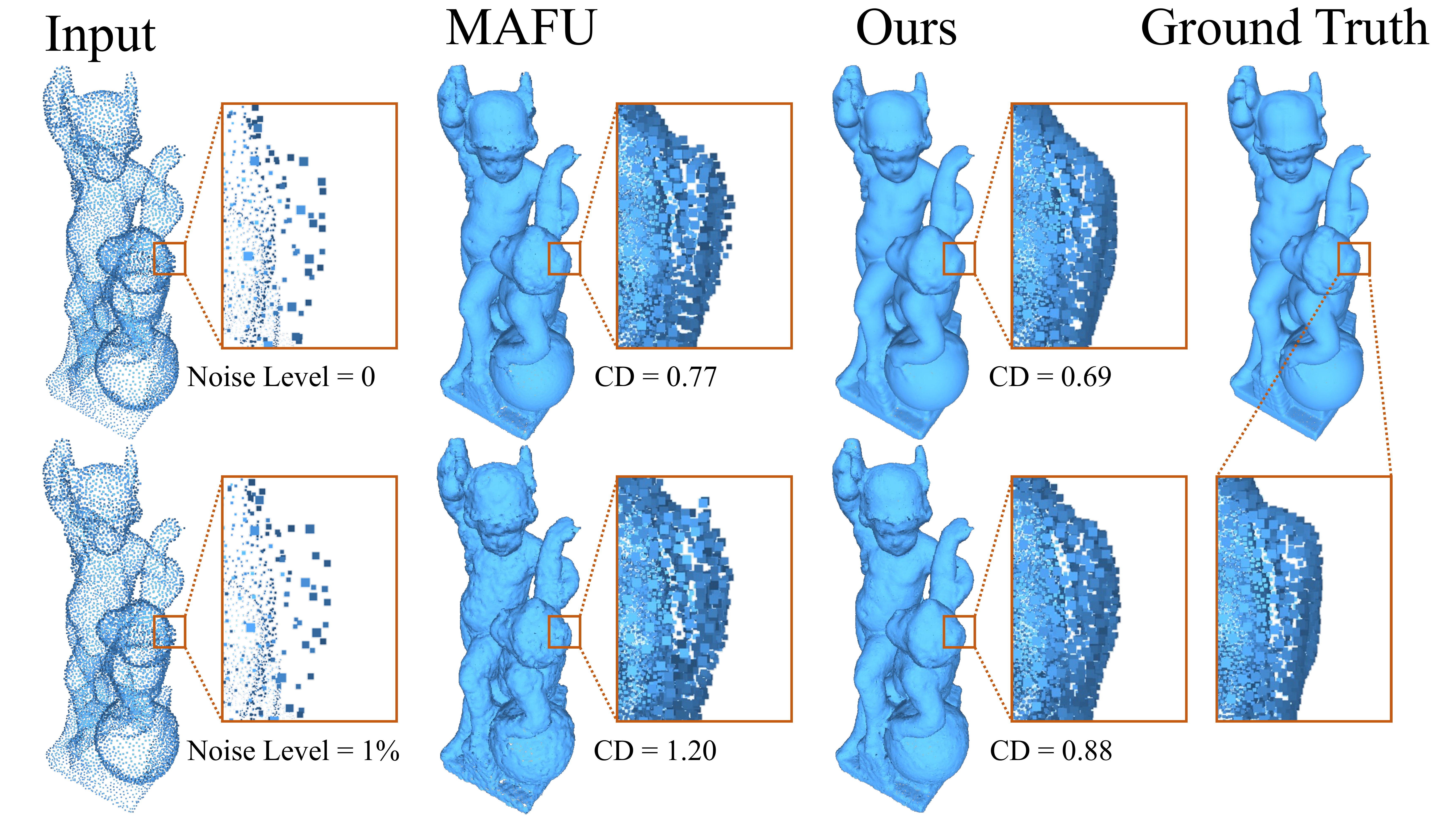}
\vspace{-4.5mm}
\caption{Comparison with MAFU \& robustness test on noise.}
\label{fig:mapu}
\vspace{-0.1mm}
\end{figure}

\begin{table}[t]
	\setlength\tabcolsep{5pt}
	\begin{center}
		\begin{tabular}{l|ccc}
			\bottomrule
			Ablation Settings                     & CD  & HD & P2F \\ \hline
		    w/o input local KNN                              & 2.49   & 9.54  & 3.03  \\ 
			w/o $\mathcal{L}_{\textrm{nor}}$                            & 0.70  & 3.46  & 0.75  \\ 
			w/o $\mathcal{L}_{\textrm{int}}$                            & 0.77  & 4.09  & 0.81  \\ 
			w/o integration                                    & 0.83   & 5.64    & 0.88      \\ 
			full model                                & \textbf{0.66}        & \textbf{3.32}           & \textbf{0.69}       \\ 
			\toprule
		\end{tabular}
	\end{center}
    \vspace{-7mm}
	\caption{Results of the ablation study, with metrics CD($\times 10^{-5}$), HD($\times 10^{-3}$), and P2F($\times 10^{-3}$).}
	\label{tab:ablation}
	\vspace{-5mm}
\end{table}

\subsection{Ablation Study}
\label{ablation_study}
We conduct ablation studies to show how each component influences the final result. Specifically, we use the Sketchfab dataset as our benchmark. We mainly design the ablation study for the local feature extractor structure, the loss terms and the integration process. The quantitative results of the ablation study experiment are shown in Tab.~\ref{tab:ablation}. In the experiment of the 1-st row, we apply a DGCNN backbone for the whole input point cloud instead of the local KNN point set. We can see the quantitative result is quite poor, which verifies that it is necessary to utilize a shared backbone for KNN point sets and limit the size of the receptive field in the feature extractor. The 2-3-th rows show different loss term settings, where we remove the $\mathcal{L}_{\textrm{nor}}$ and $\mathcal{L}_{\textrm{int}}$ terms, respectively. We can see that removing any of the two terms would make the result to be not as good as the full model. Among them, removing the $\mathcal{L}_{\textrm{int}}$ loss term will greatly reduce the algorithm performance. At last, we remove the integration process and all related algorithm settings, and show the results in the 4-th row. We can see that the result without the integration process is much worse than the full model.

\begin{table}[]
    \small
	\setlength\tabcolsep{3.5pt}
\begin{tabular}{c|c|ccc|ccc}
\bottomrule
\multirow{2}{*}{Factor} & \multirow{2}{*}{Method} & \multicolumn{3}{c|}{Sketchfab}                                 & \multicolumn{3}{c}{PU-GAN}                                     \\ \cline{3-8} 
                        &                         & \multicolumn{1}{c}{CD}     & \multicolumn{1}{c}{HD}   & P2F  & \multicolumn{1}{c}{CD}    & \multicolumn{1}{c}{HD}    & P2F   \\ \hline
\multirow{6}{*}{4x}     & PU-Net                  & \multicolumn{1}{c}{5.93}   & \multicolumn{1}{c}{4.98} & 4.51 & \multicolumn{1}{c}{23.61} & \multicolumn{1}{c}{13.91} & 10.02 \\ \cline{2-8} 
                        & PU-GAN                  & \multicolumn{1}{c}{3.30}   & \multicolumn{1}{c}{3.45} & 3.61 & \multicolumn{1}{c}{16.79} & \multicolumn{1}{c}{9.36}  & 7.04  \\ \cline{2-8} 
                        & PU-GCN                  & \multicolumn{1}{c}{2.85}   & \multicolumn{1}{c}{3.21} & 2.79 & \multicolumn{1}{c}{14.74} & \multicolumn{1}{c}{11.97} & 6.36  \\ \cline{2-8} 
                        & Dis-PU                  & \multicolumn{1}{c}{2.61}   & \multicolumn{1}{c}{3.25} & 2.67 & \multicolumn{1}{c}{13.79} & \multicolumn{1}{c}{11.83} & 7.14  \\ \cline{2-8} 
                        & PUGeo-Net               & \multicolumn{1}{c}{2.28}   & \multicolumn{1}{c}{\textbf{2.10}} & 1.04 & \multicolumn{1}{c}{11.26} & \multicolumn{1}{c}{3.54} & 2.14  \\ \cline{2-8} 
						& MAFU               & \multicolumn{1}{c}{2.26}   & \multicolumn{1}{c}{2.31} & 0.99 & \multicolumn{1}{c}{11.18} & \multicolumn{1}{c}{3.74} & 2.09  \\ \cline{2-8}
                        & Ours                    & \multicolumn{1}{c}{\textbf{2.17}}   & \multicolumn{1}{c}{2.46} & \textbf{0.93} & \multicolumn{1}{c}{\textbf{8.17}}  & \multicolumn{1}{c}{\textbf{3.08}}  & \textbf{1.59 } \\ \hline
\multirow{6}{*}{16x}    & PU-Net                  & \multicolumn{1}{c}{5.25} & \multicolumn{1}{c}{5.82} & 5.99 & \multicolumn{1}{c}{20.70} & \multicolumn{1}{c}{15.49} & 12.16 \\ \cline{2-8} 
                        & PU-GAN                  & \multicolumn{1}{c}{2.97}   & \multicolumn{1}{c}{3.99} & 3.76 & \multicolumn{1}{c}{11.89} & \multicolumn{1}{c}{10.81} & 7.48  \\ \cline{2-8} 
                        & PU-GCN                  & \multicolumn{1}{c}{2.35}   & \multicolumn{1}{c}{3.84} & 3.02 & \multicolumn{1}{c}{11.37} & \multicolumn{1}{c}{12.69} & 6.95  \\ \cline{2-8} 
                        & Dis-PU                  & \multicolumn{1}{c}{2.36}   & \multicolumn{1}{c}{3.79} & 3.31 & \multicolumn{1}{c}{12.75} & \multicolumn{1}{c}{13.65} & 8.09  \\ \cline{2-8} 
                        & PUGeo-Net               & \multicolumn{1}{c}{0.83}      & \multicolumn{1}{c}{3.50} & 0.97 & \multicolumn{1}{c}{3.58} & \multicolumn{1}{c}{7.14} & 1.94  \\ \cline{2-8} 
						& MAFU               & \multicolumn{1}{c}{0.80}      & \multicolumn{1}{c}{3.57} & 0.95 & \multicolumn{1}{c}{3.51} & \multicolumn{1}{c}{7.20} & 1.90 \\ \cline{2-8} 
                        & Ours                    & \multicolumn{1}{c}{\textbf{0.66}}   & \multicolumn{1}{c}{\textbf{3.32}} & \textbf{0.69} & \multicolumn{1}{c}{\textbf{3.35}}  & \multicolumn{1}{c}{\textbf{6.52}}  & \textbf{1.49}  \\ 
                        \toprule
\end{tabular}
\vspace{-3mm}
\caption{Results and comparisons for $4\times$ and $16\times$ upsampling, with metrics CD($\times 10^{-5}$), HD($\times 10^{-3}$), and P2F($\times 10^{-3}$).}
	\vspace{-1mm}
	\label{tab:comparison}
	\vspace{-3mm}
\end{table}

\vspace{-0.1mm}
\begin{table}[h]
	\footnotesize
	\setlength\tabcolsep{3pt}
\begin{tabular}{c|c|ccccccc}
\bottomrule
\multicolumn{2}{c|}{Method}   & \multirow{1}{*}{Net}     & \multirow{1}{*}{GAN}   & \multirow{1}{*}{GCN}  & \multirow{1}{*}{Dis}    & \multirow{1}{*}{PUGeo}    & \multirow{1}{*}{MAFU}  & \multirow{1}{*}{Ours}   \\ \hline
\multirow{2}{*}{4$\times$}     & Size (Mb)                  & 9.4   & 7.1 & \textbf{1.8} & 13.2 & 27.1 & 4.7 & 2.5 \\  
                        & Time (s)                  & \textbf{0.011}  & \textbf{0.011} & 0.012 & 0.047 & 0.014  & 0.014 & 0.015 \\ \hline
\multirow{2}{*}{16$\times$}    & Size (Mb)                  & 23.0 & 11.5 & 3.0 & 13.2 & 27.1 & 4.7 & \textbf{2.5} \\ 
                        & Time (s)                & 0.120  & 0.046 & 0.039 & 0.085 & \textbf{0.017}  & 0.018  & \textbf{0.017}  \\ 
                        \toprule
\end{tabular}
\vspace{-4mm}
\caption{Inference time (forward once with batch size $1$) and network size of methods (from left to right): PU-Net, PU-GAN, PU-GCN, Dis-PU, PUGeo-Net, MAFU, and ours.}
	\vspace{-0.1mm}
	\label{tab:time_size}
	\vspace{-5mm}
\end{table}

\subsection{Results and Comparisons}
\label{results_and_comparisons}
We compare our method with: PU-Net~\cite{YuLFCH18}, PU-GAN~\cite{LiLFCH19}, PU-GCN~\cite{QianALTG21}, Dis-PU~\cite{LiLHF21}, PUGeo-Net~\cite{QianHK020} \textcolor{black}{and MAFU~\cite{QianHKH21}}. All the comparison methods are in a discrete-to-discrete manner, and are trained with the same training set with our method. For all methods, we train with $4\times$ and $16\times$, respectively, with ground truth in the related resolution. For fairness, we share the same settings (batch size, iterations, learning rate, etc) among all methods. 

The results and comparisons of $4\times$ and $16\times$ are given in Tab.~\ref{tab:comparison}, Fig.~\ref{fig:sketchfab} \textcolor{black}{and Fig.~\ref{fig:mapu}}. Our method achieves the best performance both quantitatively and qualitatively. The results of PU-Net are messy overall. PU-GAN performs better than PU-Net but also generates strange noise and outliers. PU-GCN performs better than the above two methods and can preserve flat regions, but still produce some noise points in the feature-rich regions. We tried our best to finetune the training settings for Dis-PU~\cite{LiLHF21}. On flat regions, Dis-PU performs quite well which benefits from its disentangled refinement scheme. However, in high-curvature regions, Dis-PU can not produce high-quality result. PUGeo-Net \textcolor{black}{and MAFU} achieves better result than all the other comparison methods. However, from the zoom-in view, we can find that the distribution of \textcolor{black}{their} upsampled points is not as smooth as our results. This is because \textcolor{black}{they upsample} each patch in a discrete and independent way. From the results, we can observe that our method generates results which are globally smooth and contain rich local geometric details. We list the inference time and network size of all methods in Tab.~\ref{tab:time_size}.

\begin{figure}[t]
\centering
\includegraphics[width=0.4\textwidth]{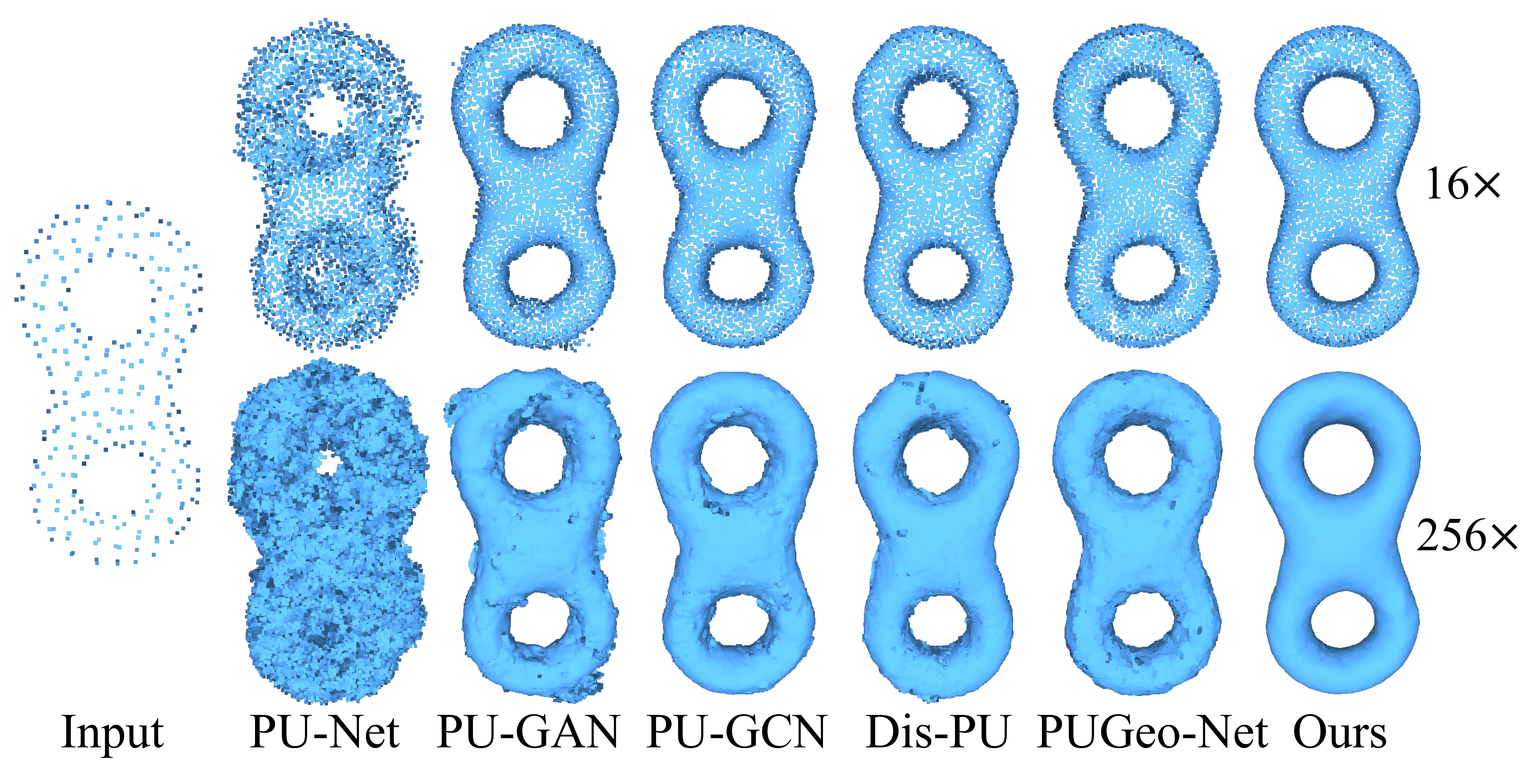}
\vspace{-5mm}
\caption{Results of upsampling with very large factors.}
\label{fig:small_large}
\vspace{-3mm}
\end{figure}


\begin{figure}[t]
\centering
\includegraphics[width=0.4\textwidth]{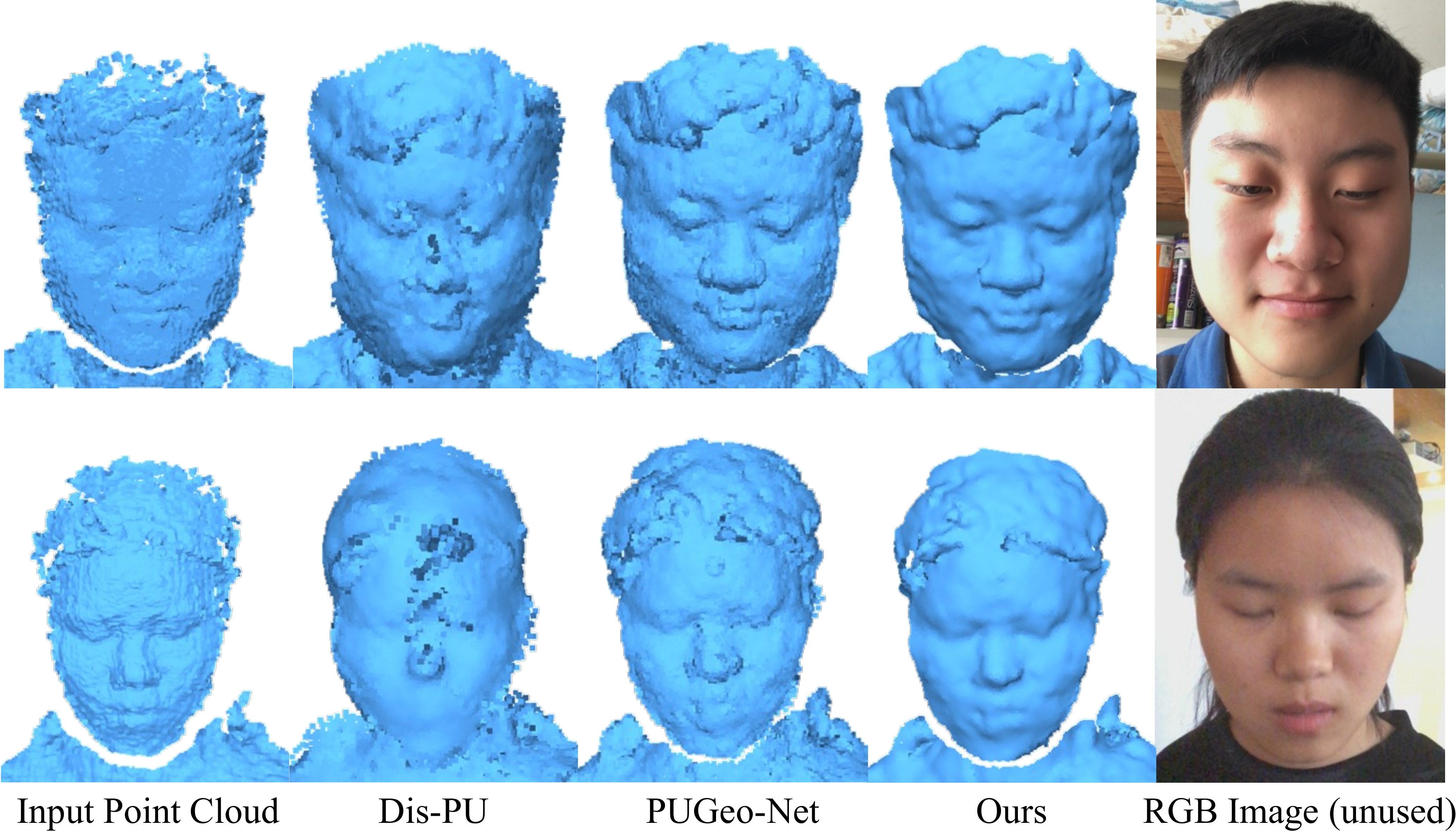}
\vspace{-3mm}
\caption{Comparisons on point cloud captured by depth camera. The RGB images are displayed for better visualization, which are not used in our method.}
\label{fig:iphone}
\vspace{-6mm}
\end{figure}

\subsection{Generalization and Robustness}
\label{generelization_and_robustness}

{\noindent \bf Generalization to Unseen Data.}
We test on another dataset which was commonly used in previous dataset, the dataset collected by PU-GAN~\cite{LiLFCH19}. Without retraining, we apply the trained models on the PU-GAN dataset again and show the visual and quantitative results. The results and comparisons are in Tab.~\ref{tab:comparison} and Fig.~\ref{fig:pugan}. We can see that our model still achieves the best performance on both of quantitative results and visualization results, proving the good generalization ability of the Neural Points representation.

{\noindent \bf Robustness to Noise.}
\textcolor{black}{We add Gaussian noise to the input, and show the results in Fig.~\ref{fig:mapu}. For the same inputs, our result keeps a good shape while the result of MAFU looks wrinkled. Our quantitative result is also better than MAFU.}

{\noindent \bf Robustness to Large Upsampling Factor.}
We explore the upsampling effect of different methods with very large upsampling factors. Notice that all the testing methods including our own method are trained with a $16 \times$ supervision signal. We adjust the $N$ in our method as $256 \cdot I$ to obtain the $256 \times$ result. For other methods, we apply the trained $16 \times$ model for two times to obtain the $256 \times$ results. The results are shown in Fig.~\ref{fig:small_large}. The visualization results of $16\times$ upsampling by different methods do not have too much difference, while the results of $256\times$ upsampling are quite different. After applying the model for two times, the results of PU-GAN, PU-GCN, Dis-PU and PUGeo-Net contain some flaws that are easy to observe. At the same time, our result still keeps in a good quality. Since Neural Points represents the surface in a continuous and resolution-free way, our result will not be affected by large sampling factors.


{\noindent \bf Robustness to Real Captured Data.}
We further compare our method with two most representative methods (Dis-PU~\cite{LiLHF21} and PUGeo-Net~\cite{QianHK020}) on real captured data. Specifically, we capture depth images of human face with the depth sensor equipped on iPhone X. Taking the point cloud converted from the depth image as input, we apply upsampling to the point clouds with different methods. As shown in Fig.~\ref{fig:iphone}, our method performs quite well, even the scanned point clouds contain lots of noise and bumpy local regions. In contrast, the results generated by Dis-PU~\cite{LiLHF21} fail to preserve the geometry features and contain some artifacts. The results of PUGeo-Net~\cite{QianHK020} preserve some geometric features of the input point cloud, but we can still observe some artifacts in their results. This test verifies the robustness and effectiveness of Neural Points to real captured data.

\vspace{-0.1mm}
\begin{figure}[h]
\centering
\includegraphics[width=0.4\textwidth]{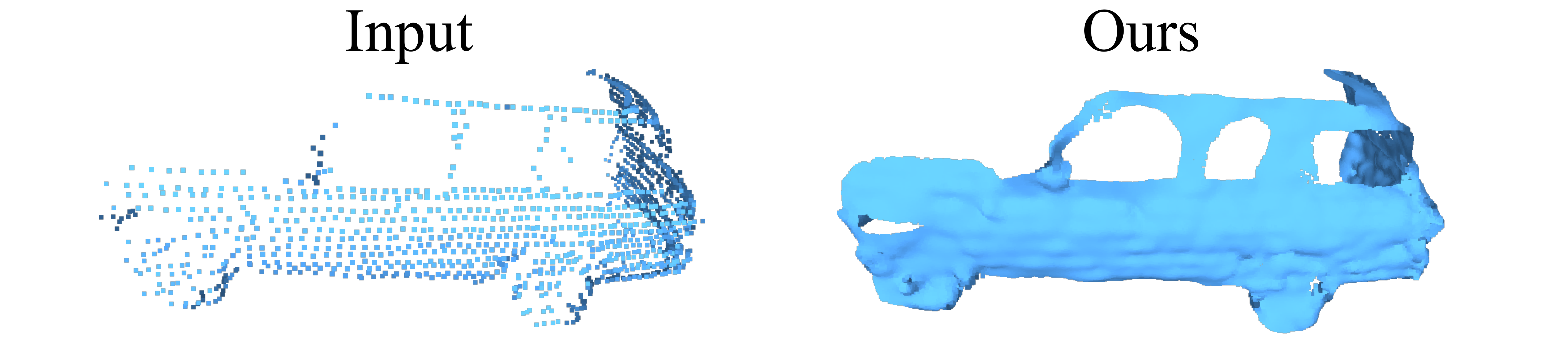}
\vspace{-3mm}
\caption{Result on LiDAR data from KITTI dataset.}
\label{fig:kitti}
\vspace{-3mm}
\end{figure}

{\noindent \bf Robustness to LiDAR-scanned Data.}
\textcolor{black}{We show the test result on LiDAR data from KITTI~\cite{GeigerLSU13} in Fig.~\ref{fig:kitti}, and we can also achieve reasonable upsampling result.}
\section{Conclusion\&Limitation}
\label{Conclusion}

We proposed Neural Points, \textcolor{black}{a novel point cloud representation for arbitrary upsampling}, where each point represents a local continuous geometric shape via neural fields instead of only a position or a local plane in the 3D space. Neural Points can \textcolor{black}{involve more shape information} and thus have a stronger representation ability than the traditional point cloud. We trained Neural Points with surface containing rich geometric details, such that the trained model has enough expression ability for various shapes. Specifically, we extracted deep local features and constructed neural fields through the local isomorphism between the 2D parametric domain and the 3D local patch. The final global continuous surface is obtained by integrating the neural fields. The powerful representation ability, robustness and generalization ability of Neural Points have been verified by extensive experiments. \textcolor{black}{The great performance on the upsampling task further verified its nice properties}.

Currently, we only utilize the local geometry shape to train Neural Points, which might limit its applications to more broad areas. In the future, we plan to utilize other modalities like corresponding color images and textures, or global semantic structure to further improve its representation ability.

\small {\noindent{\bf Acknowledgement} \textcolor{black}{This research was supported by National Natural Science Foundation of China (No. 62122071), the Youth Innovation Promotion Association CAS (No. 2018495), ``the Fundamental Research Funds for the Central Universities''(No. WK3470000021).}}

{\small
\bibliographystyle{ieee_fullname}
\bibliography{egbib}
}

\appendix

\twocolumn[
\centering
\Large
\textbf{Neural Points: Point Cloud Representation with Neural Fields for Arbitrary Upsampling} \\
\vspace{0.5em}Supplementary Material \\
\vspace{1.0em}
] %


\begin{strip}\centering
	\includegraphics[width=\textwidth]{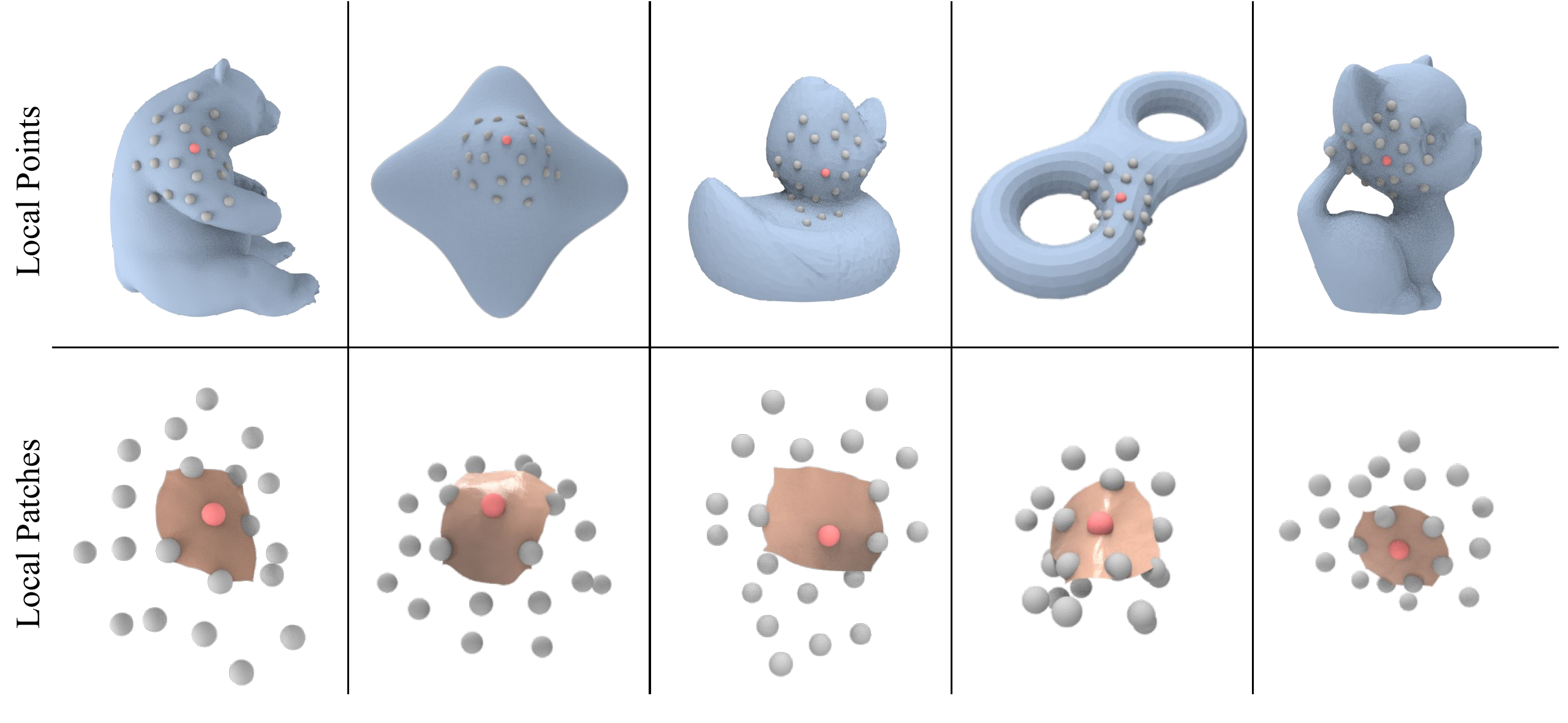}
	\captionof{figure}{The visualization of the local patches of our Neural Points representation. We show some examples of local point clouds in the first row, where the center point is colored red. To clearly show the underlying local shape, we also show the underlying surface from which we extract the local points. The corresponding local patches generated from Neural Fields in our algorithm are shown in the second row. Each column shows different local parts in our testing set.} \label{fig:patches}
\end{strip}

This supplementary material shows more results and analysis which are not included in the paper due to limited space, including the visualization of the local Neural Fields, results with more sampling factors, and the qualitative results of ablation study.

\section{Visualization of Neural Field Patches}
To clearly show the Neural Fields of each point, we visualize some local patches generated from the Neural Fields in Fig.~\ref{fig:patches}. We show the local point cloud where we extract local features in the first row and then show their corresponding local Neural Field patches in the second row. The center points are rendered in red. In the first row, to clearly show the local underlying shape, we show the underlying surface from which we extract the local points. In the second row, we zoom in the local parts and show the generated Neural Field patches. We can see that the generated patches can cover a local area smoothly and fit well with the shape of the local point cloud. Each column shows different local parts extracted from the models in our testing set. For different local shapes, the Neural Field can obtain satisfactory local patches, which contributes a lot to the overall expressive ability of the Neural Points representation.

\begin{figure*}[t]
\centering
\includegraphics[width=1.0\textwidth]{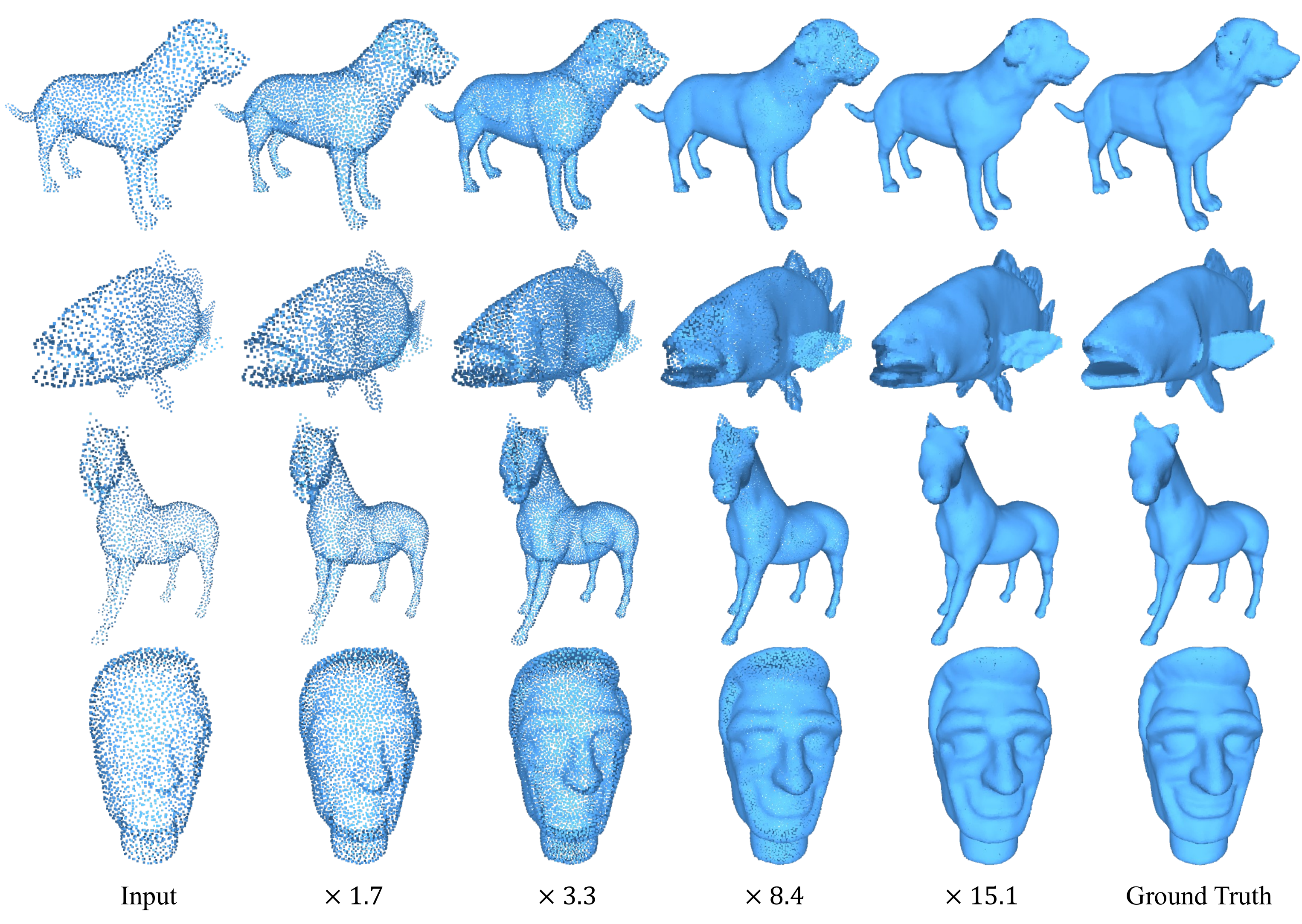}
\caption{The results with arbitrary sampling factors including non-integer factors.}
\label{fig:arbitrary}
\vspace{-5mm}
\end{figure*}

\begin{figure*}[t]
\centering
\includegraphics[width=1.0\textwidth]{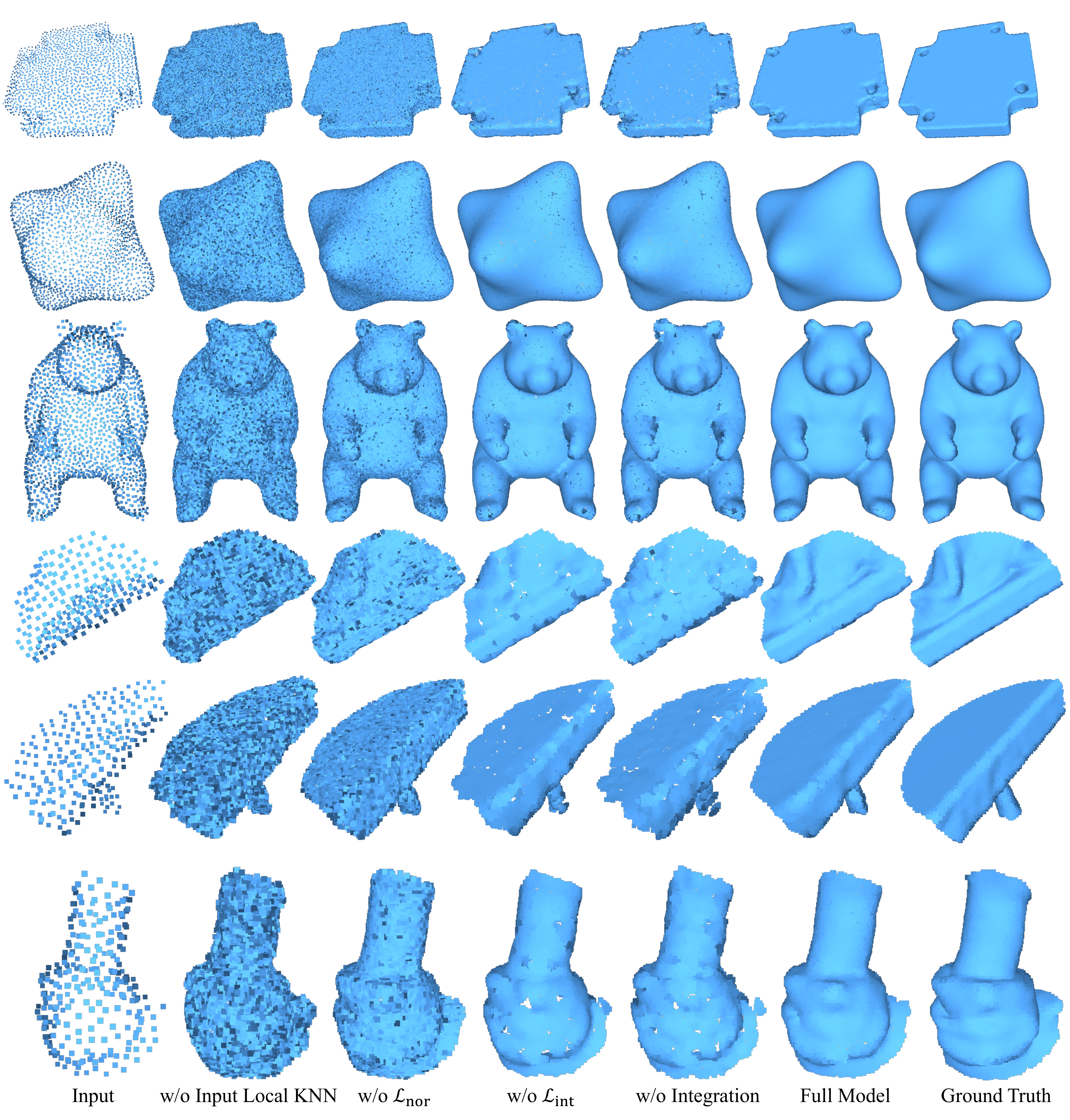}
\caption{The visualization results of our ablation study experiments.}
\label{fig:ablation}
\vspace{-5mm}
\end{figure*}

\section{Results with More Sampling Factors}
As stated in the paper, our trained model can be applied with arbitrary sampling factors. In this section, we show results with more sampling factors, and non-integer factors are also supported in our model. In Fig.~\ref{fig:arbitrary}, we show some examples. Specifically, we upsample the input points with factors $1.7$, $3.3$, $8.4$, and $15.1$. We can see that our model can generate good results for various factors within a large range. Due to the limited space, we only show these factors, but our model also supports other factors. We strongly recommend watching our video, where we show the results with continuously varying factors.

\section{Visualization of Ablation Study}
Except for the quantitative experiment in the paper, we also show the visualization results of the ablation study in Fig.~\ref{fig:ablation}. If we apply the backbone for the whole point cloud instead of the local KNN point set, the training effect would be quite poor, and the visual results look messy. If we remove the loss term for normal $\mathcal{L}_{\textrm{nor}}$, the training effect would not be as good as the full model. The visual results look not smooth enough. If we remove the integration term $\mathcal{L}_{\textrm{int}}$, the results would contain some holes. Further, when we remove all the processes related to integration, the results not only have holes but also suffer from some splicing dislocations. Compared to these algorithm settings, our full model can generate satisfying results.

\end{document}